\documentclass[final]{cvpr}

\usepackage{times}
\usepackage{epsfig}
\usepackage{graphicx}
\usepackage{amsmath}
\usepackage{amssymb}
\usepackage{microtype}
\usepackage{xcolor}
\usepackage{booktabs}
\usepackage{multirow}

\usepackage[pagebackref=true,breaklinks=true,colorlinks,bookmarks=false]{hyperref}

\def\cvprPaperID{1451} 
\def\confYear{CVPR 2021}
\usepackage{gensymb}
\usepackage{makecell}

\newenvironment{packed_enum}{
\begin{enumerate}
  \setlength{\itemsep}{1pt}
  \setlength{\parskip}{2pt}
  \setlength{\parsep}{0pt}
}{\end{enumerate}}

\definecolor{darkgreen}{rgb}{0.02,0.6,0.02}
\iffalse
\newcommand{\yw}[1]{}
\newcommand{\bc}[1]{}
\newcommand{\brian}[1]{}
\newcommand{\al}[1]{}
\newcommand{\ns}[1]{}
\newcommand{\steve}[1]{}
\newcommand{\rt}[1]{}
\newcommand{\jw}[1]{}
\else
\newcommand{\yw}[1]{\textcolor{cyan}{[\textbf{YW says}: #1]}}
\newcommand{\brian}[1]{\textcolor{red}{[\textbf{BC says}: #1]}}
\newcommand{\al}[1]{\textcolor{blue}{[\textbf{AL says}: #1]}}
\newcommand{\ns}[1]{\textcolor{purple}{[\textbf{NS says}: #1]}}
\newcommand{\steve}[1]{\textcolor{red}{[\textbf{SS says}: #1]}}
\newcommand{\rt}[1]{\textcolor{magenta}{[\textbf{RT says}: #1]}}
\newcommand{\jw}[1]{\textcolor{darkgreen}{[\textbf{JW says}: #1]}}
\fi

\iffalse
\newcommand{\addtext}[1]{\textcolor{cyan}{#1}}
\else
\newcommand{\addtext}[1]{#1}
\fi

\begin{document}

\title{Repopulating Street Scenes}
%

\author{Yifan Wang\textsuperscript{1}\thanks{This work was done while Yifan was an intern at Google.}
\hspace{25pt}
Andrew Liu\textsuperscript{2}
\hspace{25pt}
Richard Tucker\textsuperscript{2}
\hspace{25pt}
Jiajun Wu\textsuperscript{3}\\
Brian L. Curless\textsuperscript{1,2}
\hspace{25pt}
Steven M. Seitz\textsuperscript{1,2}
\hspace{25pt}
Noah Snavely\textsuperscript{2}\\
$^{1}$University of Washington
\qquad
$^{2}$Google Research
\qquad
$^{3}$Stanford University
}

\maketitle

\begin{abstract}
We present a framework for automatically 
reconfiguring images of street scenes by populating, depopulating, or repopulating them with objects such as pedestrians or vehicles.
Applications of this method include anonymizing images to enhance privacy, generating data augmentations for perception tasks like autonomous driving, and composing scenes to achieve a certain ambiance, such as empty streets in the early morning.
At a technical level, our work has three primary contributions: (1) a method for clearing images of objects, (2) a method for estimating sun direction from a single image, and (3) a way to compose objects in scenes that respects scene geometry and illumination.
Each component is learned from data with minimal ground truth annotations, by making creative use of large-numbers of short image bursts of street scenes.
We demonstrate convincing results on a range of street scenes and illustrate potential applications.
\end{abstract}

\section{Introduction}

Websites such as Google Street View
enable users to explore places around the world through street-level imagery. 
These sites can provide a rich sense of what different locales---neighborhoods, parks, tourist sites, etc---are really like. 
However, the imagery provided by such sites also has key limitations. A given image might be full of cars and pedestrians, making it difficult to observe the environment. Alternatively, a user might want to see how a scene appears at a certain time of day, e.g., lunchtime, but only have access to a morning image.
And, importantly, the fact that the imagery records real people and vehicles
may require anonymization efforts
to protect privacy, e.g., by blurring faces and license plates~\cite{largescale2009frome,gsv2010anguelov} or by removing pedestrians from images by leveraging multiple views~\cite{flores2010removing}.

We propose learning-based tools that mitigate these limitations by removing objects from a scene and then repopulating that scene with, for instance, anonymized images and vehicles. Our method could thus be used to enhance privacy of imagery, while also increase flexibility to compose new scenarios (e.g., an empty street or lunchtime scene). These capabilities could also be useful in other applications, such as automatic generation of novel scene configurations as a way to augment data for training autonomous driving---especially emergency scenarios that might be rare in real data. In order for such reconfigured images to look realistic, they must respect the illumination and geometry of the underlying scene. Our learning-based method takes such factors into account.

\begin{figure*}[t!]
    \centering
    \includegraphics[width=\textwidth]{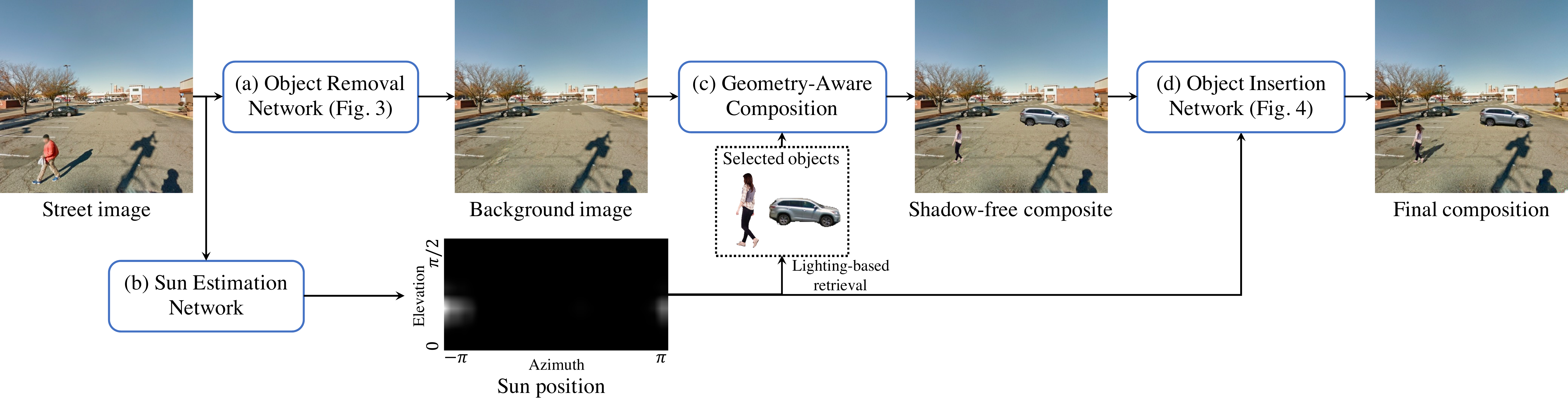}
    \caption{Our reconfiguration pipeline has four major components: (1) a removal network that learns to remove existing objects and their shadows, (2) a sun estimation network that learns to predict sun position from an image, (3) a method to scale and insert newly inserted objects with correct occlusion ordering, and (4) an insertion network that learns to cast shadows. Given a street image, our method first removes selected objects, then selects new object that matches the lighting of the scene, composes them with correct scale and occlusion order into the background image, and synthesizes shadows for inserted objects.}
    \label{fig:pipeline}
    \vspace{-12pt}
\end{figure*}

As shown in Fig.~\ref{fig:pipeline}, our framework takes as input a single street image, and can realistically remove all objects and generate repopulated images. To remove objects, we use nearby patches to inpaint not only the objects themselves, but also the shadows they cast onto the scene. To repopulate with new objects, our framework can automatically select objects that match the lighting of the scene, and compose them into the scene with proper scale, occlusion, and cast shadows consistent with the scene's geometry and lighting.

Our method consists of four main components:
\begin{packed_enum}
    \item a \emph{removal} network that removes all existing objects (cars, pedestrians, and bicycles) -- along with their shadows -- from a street image and realistically fills the resulting holes, rendering an empty version of the scene;
    \item a \emph{sun estimation} network that takes a street image, and estimates a dominant lighting (sun) direction; 
    \item a method to compose the inserted object into the scene with proper scale, occlusion based on its placement in the scene; and
    \item an \emph{insertion} network that takes a segmented object, e.g., drawn from an anonymized collection, and inserts it into a scene, generating a realistic shadow.
\end{packed_enum}
Our method yields realistic results that improve upon prior work. In particular, unlike standard methods such as image inpainting~\cite{liu2018partial,yu2018generative,yu2019free,yi2020contextual}, our approach can realistically remove and render object shadows. Instead of learning to cast shadows into one specific scene captured with a long video sequence~\cite{wang2020people}, our approach learns from short image bursts, then generalizes to 
any single image of a street scene.


Our three networks are learned in a novel way that involves observing large numbers of street scenes, with no manual annotations required. In particular, we leverage large numbers of short timelapse image sequences gathered from Google Street View. In these timelapses, objects such as pedestrians, bicyclists, and cars are generally in motion, enabling us to estimate a ground truth ``clean plate'' image devoid of moving objects (and their shadows) by computing a
median image. Pairs of input and clean plate frames thus give us ground truth images with and without objects, which we use as the basis for training. We show that this 
data, along with metadata such as camera pose and time of day, are sufficient supervision for our task.

Our work is motivated by goals such as improving visualizations of scenes and enhance image privacy. We demonstrated two potential applications using our framework: (1) emptying the city by removing all objects within an image and (2) repopulating the scenes with anonymized people. These applications are designed to enhance the privacy of a street image while preserving the realism. However, any deployment of our methods in a real-world setting would need careful attention to responsible design decisions. Such considerations could include clearly watermarking any user-facing image that has been recomposed, and matching the distribution of anonymized people composed into a scene to the underlying demographics of that location.



\section{Related Work}
Our work is related to prior work on object removal, object insertion, and lighting estimation.

\medskip
\noindent \textbf{Object removal.}
Prior work on object removal falls mainly into two groups: (1) image inpainting methods 
and (2) methods for detecting and removing object shadows.

Recently, deep learning and GAN-based approaches have emerged as a 
leading paradigm for image inpainting. Liu \etal~\cite{liu2018partial} inpaint irregular holes with partial convolutions that are masked and re-normalized to be conditioned on valid pixels. Gated convolutions~\cite{yu2019free} generalize such partial convolutions by providing a learnable dynamic feature selection mechanism for each channel and at each spatial location. Contextual attention~\cite{yu2018generative,yi2020contextual} allows for long-range spatial dependencies, 
allowing pixels to be borrowed from distant locations to fill missing regions. Shadows have different forms, e.g., hard shadows, soft shadows, partially occluded shadows, etc. Hole-filling based methods have trouble determining what pixels to inpaint in different shadow scenarios.


Deep learning methods
have also been applied to 
shadow removal. Qu \etal \cite{qu2017deshadownet} extract 
features from multiple views and aggregate them to recover shadow-free images. \addtext{Wang \etal \cite{wang2018stacked} and Hu \etal \cite{hu2019mask} use GANs for shadow removal}, while recently Le \etal \cite{le2019shadow} proposed a two-network model to learn shadow model parameters and shadow mattes. However, these methods only inpaint shadow regions. Realistic object removal involves removing both the object and its shadow, as handled by our method. 

Other work has sought to remove pedestrians from street scenes by leveraging multiple views~\cite{flores2010removing}. In our case, we operate on just a single view, and can also recompose new people into scenes. Finally, face replacement~\cite{blanz2004exchanging} has been considered for realism-preserving privacy enhancement tool~\cite{bitouk2008face}. Our work considers whole people, and not just faces.

\medskip
\noindent \textbf{Object insertion.}
Early methods for object insertion include Poisson blending \cite{perez2003poisson}, which can produce seamless object boundaries, but can also result in illumination and color mismatches between the object and the target scene. 
Lalonde \etal proposed \emph{Photo Clip Art}, which inserts new objects into existing photographs by first querying a large dataset of cutouts for compatible objects~\cite{lalonde2007photo}. Other methods match the color, brightness, and styles of inserted objects to harmoniously embed them into background images \cite{lin2018st,azadi2018compositional,zhan2019spatial}. However, a realistic insertion should also consider an object's effect on the background (including shadows).

Some methods insert a 3D object by rendering it into in an image. Karsch \etal demonstrate convincing 
object insertions via inverse rendering models derived from geometric inference \cite{karsch2011rendering} or via single-image depth reconstruction \cite{karsch2014automatic}. Other work renders inserted objects with estimated HDR environment lighting maps \cite{hold2017deep,hold2019deep}. 
Chuang \etal synthesize shadows for inserted objects via a shadow displacement map~\cite{chuang2003shadow}. However, these approaches essentially require a full 3D model of either the inserted object or the scene.
\addtext{Liu \etal \cite{Liu_2020_CVPR} focus on single light source scenes containing hard shadows, whereas our method can handle scenes with soft shadows and spatially varying lighting.}
Recently, Wang \etal \cite{wang2020people} proposed a data-driven method that takes a long video of a scene and learns to synthesize shadows for inserted 2D cutout objects. Our method learns from short bursts of images, and can synthesize shadows for 2D cutouts given a single image of a new scene at test time.

\medskip
\noindent \textbf{Lighting estimation.}
To capture illumination Debevec \cite{debevec2008rendering} captures HDR environment maps via bracketed exposures of a chrome ball. Subsequent methods \cite{calian2018faces,gardner2017learning,hold2017deep,lalonde2009estimating,legendre2019deeplight,sengupta2019neural} 
use machine learning to predict HDR environment maps from single indoor or outdoor images. However, a single environment map is insufficient for compositing cut-out objects into a large captured scene, because different lighting effects will apply depending on, for instance, whether the object is placed in a sunlit area or a shadowed one.
In our work, we do not explicitly estimate lighting for each scene, but instead use a rendering network that implicitly learns to generate shadows appropriate for the object location.

Outdoor illumination is primarily determined by the sun position and the weather conditions. Recent works \cite{ma2017find,hold2017deep,hold2019deep,liu2020learning} use data-driven methods to estimate the sun azimuth angle from a single outdoor image. Our work follows this trend and estimates a full 2D sun angle. We find that estimating the sun position 
aids in 
synthesizing plausible shadows in different weather and lighting scenarios.

\section{Approach}
Given an image of a street scene, our goal is to recompose the objects (\eg, cars and pedestrians) in the scene by first removing the existing objects, and then optionally composing one or more new objects into the scene. These stages must respect the illumination in the scene---in particular, shadows (both their removal and insertion) are critical elements that are difficult to handle realistically in prior work. 


Our automatic pipeline for addressing this problem has four major components, as shown in Fig.~\ref{fig:pipeline}: (1) a removal network that learns to remove existing objects and their shadows, (2) a lighting estimation network that learns to predict sun position from an image, which helps identify compatible objects for insertion and is used to create better insertion composites, (3) a method to scale the inserted object properly with correct occlusion order based on its placement in the scene, and (4) an insertion network that learns to cast shadows for newly inserted objects.
Given a street image, we first use Mask R-CNN~\cite{wu2019detectron2} to segment existing people and cars, then use that as a mask for the object removal stage. The object removal network completely removes those objects (and their shadows), yielding a background image. The sun estimation network is used to select new objects that match the scene's lighting. Selected objects are then composed into the background image with correct scale and occlusion order to get a shadow-free composite. Finally, our insertion network takes the shadow-free composite, synthesizes shadows, and outputs the final composite. 


\begin{figure}[t!]
    \centering
    \includegraphics[width=\linewidth]{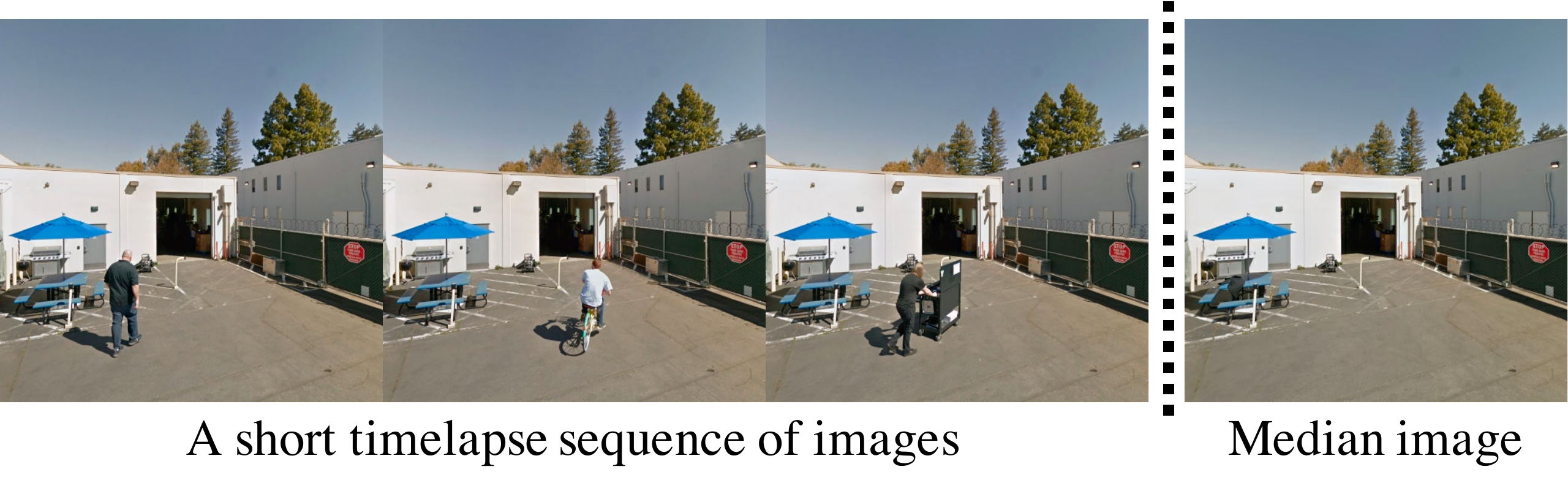}
    \caption{Our dataset consists of short image bursts, \ie, short timelapse sequences of images captured over several seconds. We compute the median of each timelapse image stack to produce a “clean plate” background image free of objects and shadows.}
    \label{fig:data}
    \vspace{-12pt}
\end{figure}

\subsection{Data}
\label{sec:data}
We train our networks in a novel way by using a dataset of image bursts, i.e., short timelapse image sequences captured over several seconds. As shown in Fig.~\ref{fig:data}, we compute the median of each timelapse image stack to produce a ``clean plate'' background image free of (moving) objects such as people and their shadows. We also know location and time of day for each timelapse, from which we derive the sun position. (We do not use weather data; the sun position is noted regardless of cloud cover.) Images with and without moving objects, and corresponding sun positions serve as ground truth supervision for our object removal, sun prediction, and object insertion networks. At test time, our pipeline takes in a single street image and can remove and repopulate the objects within. We now describe the four components of our pipeline.



\subsection{Removal Network}
\label{sec:remove}
Object removal is a challenging task that involves generating new content in holes left by removed objects, such that the new image is realistic and semantically correct.
Given a mask indicating the objects to be removed, standard inpainting methods \cite{liu2018partial,yu2018generative,yu2019free,yi2020contextual} only fill masked regions, leaving behind shadows. Our goal is to remove both objects and their shadows. We propose a deep network that, given an image and an object mask, constructs a new mask that includes the object and its shadow, then inpaints the region inside this mask. Inspired by PatchMatch-based inpainting~\cite{he2012statistics} and appearance flow~\cite{ren2019structureflow}, our method predicts a flow map that uses nearby patches' features to inpaint the masked region.

\medskip \noindent \textbf{Algorithm.}
Standard inpainting methods operate on an image and a binary mask designating where to inpaint. In our case, the network also automatically detects shadow regions belonging to masked objects. Different objects have different shadow shapes---for example, people can have long, thin, complex shadows, while cars tend to have larger, simpler shadows. 
Hence, rather than taking a binary mask, our network receives an image and a class mask produced by Mask R-CNN~\cite{wu2019detectron2}. 
This class mask encodes the object category of each pixel with distinct values normalized to $[0, 1]$. 

\begin{figure}[t]
    \centering
    \includegraphics[width=\linewidth]{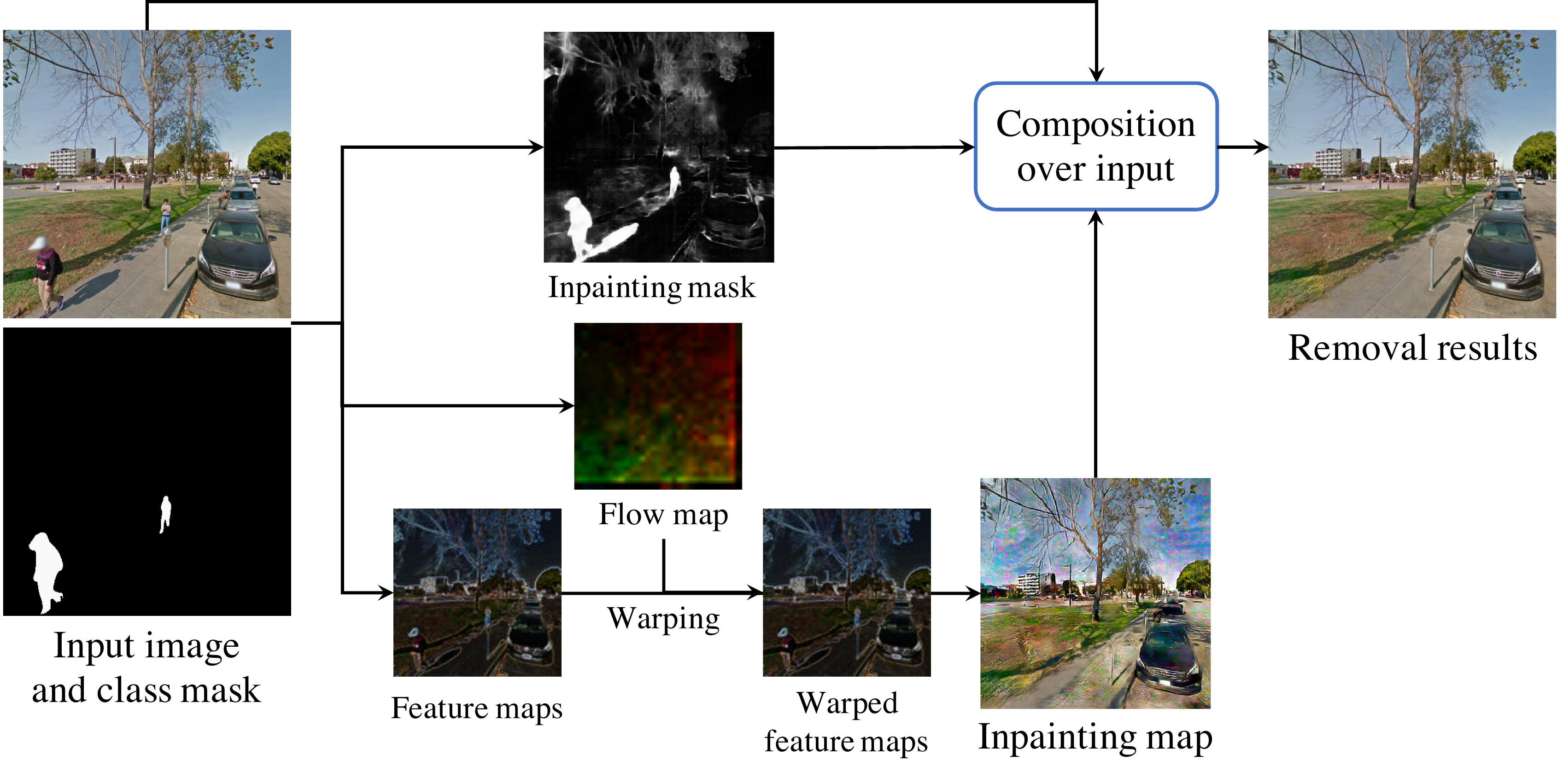}
    \caption{The generator of the removal network takes an input image $I$ and a class mask as input, and outputs an inpainting mask $M_{\rm inp}$ and an inpainting map $I_{\rm inp}$. We synthesize the removal image $I_{\rm remove} = I_{\rm inp} \odot M_{\rm inp} + I \odot (1 - M_{\rm inp})$.}
    \label{fig:remove_network}
    \vspace{-12pt}
\end{figure}

Fig.~\ref{fig:remove_network} shows the removal network architecture. We feed the input image $I$ and its class mask through four downsampling layers followed by three different branches of residual blocks. The first branch predicts a full inpainting mask $M_{\rm inp}$, including the object and its shadow; the second predicts a warping flow map $F_{\rm warp}$; and the third encodes the image as a high-dimensional inpainting feature map $F_{\rm inp}$. The feature map is then warped by the predicted flow map $F_{\rm warp}$ and fed into four upsampling layers to produce an inpainting image $I_{\rm inp}$. The final removal image is then computed as
\begin{equation}
I_{\rm remove} = I_{\rm inp} \odot M_{\rm inp} + I \odot (1 - M_{\rm inp}).
\end{equation}
The feature warping layer uses the high-dimensional features of nearby patches to inpaint the missing area. We found that street scenes often have highly repetitive structures---building facades, fences, road markings, etc.---and the feature warping layer works well in these situations.

\subsection{Sun Estimation Network}
\label{sec:sun_pos}
Lighting is key to realistic image composition. An object lit from the left composed into a scene lit from the right will likely look unrealistic. Many traditional lighting estimation methods reconstruct an environment map~\cite{calian2018faces,gardner2017learning,hold2017deep,lalonde2009estimating,legendre2019deeplight,sengupta2019neural}. 
However, a single environment map is insufficient for compositing 
objects into the scene, because different lighting effects will apply depending on object placement, 
e.g., whether the object is the shade or lit by the sun. In our work, we do not explicitly estimate scene illumination, but instead predict the sun
position with a deep network and use the result
to help 
synthesize plausible shadows. 
Further, we apply the same network 
to choose objects with similar sun position to be inserted. 

\begin{figure}[t]
    \centering
    \includegraphics[width=\linewidth]{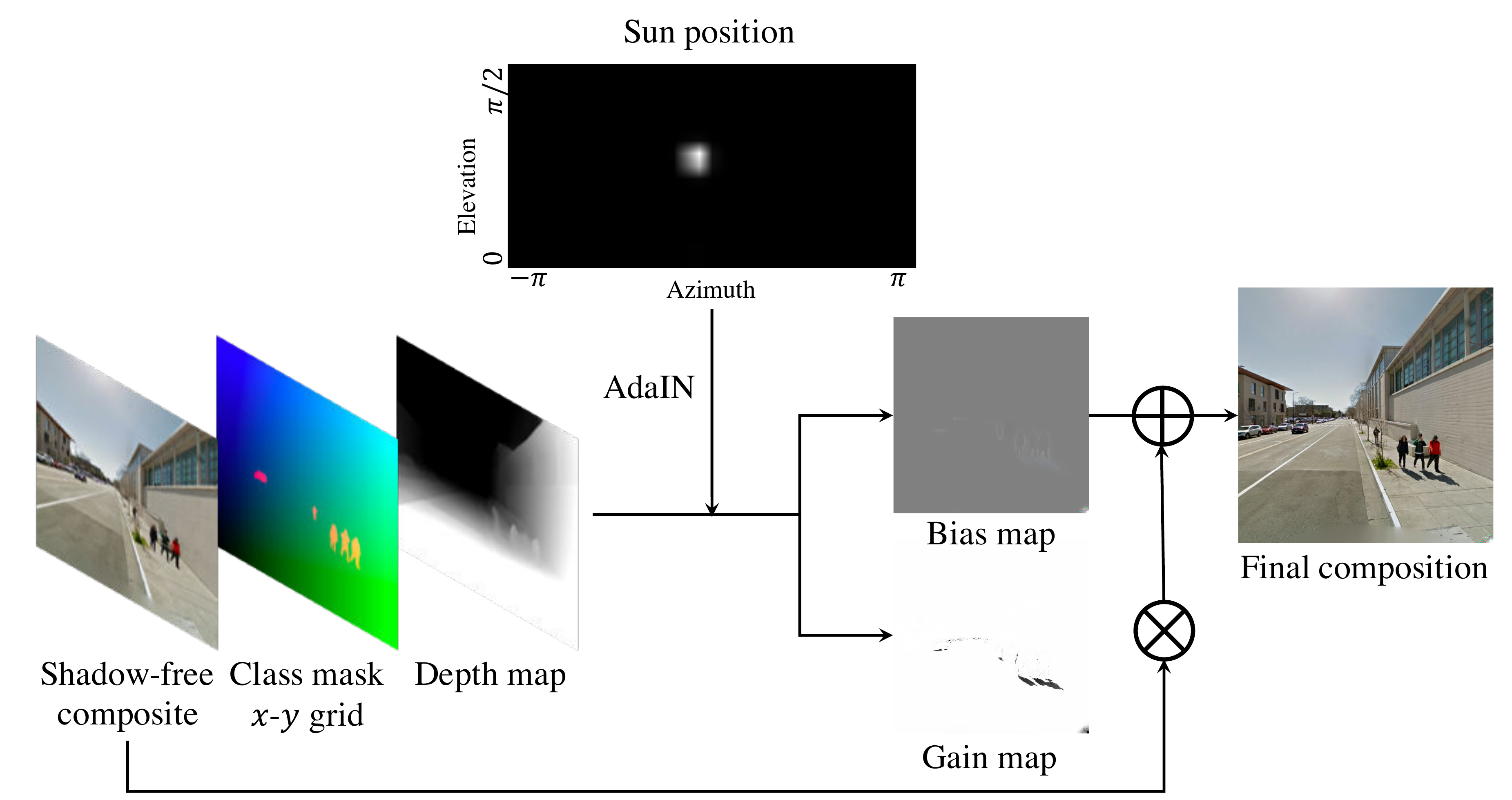}
    \caption{The generator of the insertion network takes as input a shadow-free composite image $I_{\rm comp}$, a class mask, an $x$-$y$ grid map, a depth map, and the predicted sun position distribution, and outputs a scalar gain image $G$ and color bias image $B$. Given the shadow-free composite image $I_{\rm comp}$, we synthesize the final image $I_{\rm{final}} = G \odot I_{\rm{comp}} + B$.}
    \label{fig:insert_network}
    \vspace{-12pt}
\end{figure}

\medskip \noindent \textbf{Algorithm.}
Rather than regressing an image to sun azimuth and elevation, we treat this as a classification problem and predict a distribution over discretized sun angles. 
We divide the range of azimuth angles $[0, 2\pi)$ into 32 bins and elevation angles $[0, \pi/2]$ into 16 bins. We use a network similar to ResNet50 \cite{he2016deep}, replacing the last fully connected layer with two, one for azimuth and one for elevation. We train this network using ground truth sun positions as supervision via a cross-entropy loss. In Fig.~\ref{fig:pipeline}~and~\ref{fig:insert_network}, we visualize estimated sun position as a 2D distribution formed by the outer product of azimuth and elevation distribution vectors.

\subsection{Scene Geometry for Occlusion and Scale}
\label{sec:geometry}

When composed into a scene, a new object should be scaled properly according to its 3D scene position, and should have correct occlusion relationships with other scene structures.
To that end, we desire accurate depth estimates for both the target scene and source object, 
and propose a method to robustly estimate depth for the scene and object jointly. 
\addtext{Our method, unlike~\cite{zhu2020single}, also reasons about occlusion ordering for inserted objects.}
Here we take people as an example of inserted objects, but our method also works on other objects, including cars, bikes, and buses. We make three assumptions: (1) the sidewalk and road regions in the image can be well-approximated by a single plane; (2) there is at least one person present; and (3) people are roughly the same height in 3D. If the second assumption is not met, the user can manually adjust the height scale. The third assumption, while not universally true, facilitates depth estimation by treating individual height difference as Gaussian noise.

\medskip \noindent \textbf{Algorithm.}
As described in~\cite{wang2020people} (Eq.~7), any object’s bottom middle point $(x, y)$ and height $h$ follow a linear relationship:
\begin{equation}
    a'x + b'y + c' = h
    \label{eqn:eccv_height}
\end{equation}
Also under perspective projection, the object's height $h$ is up to a scale factor $k$ with its disparity $1/Z$:
\begin{equation}
    h = k \cdot \frac{1}{Z}
    \label{eqn:world_height}
\end{equation}
Combining Eq.~\ref{eqn:eccv_height}~and~\ref{eqn:world_height}, we have a linear relation between pixel coordinates and the disparity $1/Z$:
\begin{equation}
    a'x + b'y + c' = \frac{1}{Z}
    \label{eqn:camera_depth}
\end{equation}

Given the input image, we first use DeepLab \cite{deeplabv3plus2018} to segment pixels belonging to sidewalk and road. We then use MiDaS \cite{Ranftl2020} to predict a depth map for the scene. MiDaS predicts the disparity map $\hat{D}$ up to a global scale and shift. Therefore, the linear relationship in Eq.~\ref{eqn:camera_depth} still holds. 
After collecting all 2D road/sidewalk pixels $(x_i, y_i)$ and their disparities $\hat{d}_i$, we use least squares to solve for $(a', b', c')$ in Eq.~\ref{eqn:camera_depth}. Finally, we solve for the scale factor $k$ in Eq.~\ref{eqn:world_height} using existing objects and their observed 2D heights. If there is no object in the scene, a user can manually set the scale factor.

When inserting a new object into the scene at 2D position $(x, y)$, we apply Eq.~\ref{eqn:camera_depth} and Eq.~\ref{eqn:world_height} to estimate its disparity $d$ and height $h$, and resize the inserted object accordingly. We then resolve occlusion order by comparing the object's disparity $d$ and the scene's disparity $\hat{D}$ from MiDaS. Pixels with larger disparity than $d$ are foreground and will occlude the object, and pixels with smaller disparity than $d$ will be occluded by the object.

\subsection{Insertion Network}
\label{sec:insert}
Shadows are one of the most interesting and complex ways in which 
objects interact with a scene. As with the removal network, predicting shadows for inserted objects is challenging, as their shapes and locations depend on sun position, weather, and the shape of both the object casting the shadow and the scene receiving it. Furthermore, unlike other lighting effects, shadows are not always additive, as a surface already in shadow does not darken further when a second shadow is cast on it with respect to the same light source. We propose to use observations of objects in the scene along with the scene's predicted geometry and lighting to recover these shadowing effects, using a deep network to learn how objects cast shadows depending on their shape and scene placement. Unlike the work of Wang \etal, which is trained on a long video of a scene and can only insert objects within that same scene~\cite{wang2020people}, our method learns from a database of short image bursts, and can then be applied to a single, unseen image at test time.

\medskip
\noindent \textbf{Algorithm.}
Our insertion network takes as input a shadow-free composite image $I_{\rm{comp}}$ (where the desired object is simply copy-pasted into the scene). 
As with the removal network, we consider that shadow effects vary significantly across object categories, and we also provide the class mask introduced in Sec.~\ref{sec:remove} as input. In addition, because shadows depend on scene geometry and illumination, we use MiDaS~\cite{Ranftl2020} to predict a depth map for the shadow-free image, and feed this to the insertion network, along with 
the sun position distribution from the sun estimation network. Finally, following~\cite{wang2020people}, we feed a $x$-$y$ grid map to the network to help stabilize training. 
As shown in Fig.~\ref{fig:insert_network}, $I_{\rm{comp}}$, the class mask, $x$-$y$ grid map, and depth map are concatenated and passed through four downsampling layers then five residual blocks. The sun azimuth and elevation 
vectors are concatenated, passed through four MLP layers and fused into five residual blocks via AdaIN~\cite{huang2017arbitrary}. Finally, following~\cite{wang2020people}, two different upsampling layers generate a scalar gain image $G$ and color bias image $B$. The final image is computed as
\begin{equation}
  I_{\rm{final}} = G \odot I_{\rm{comp}} + B.
\end{equation}

\section{Evaluation}
In this section, we introduce our collected datasets, then evaluate our entire pipeline and its individual components.

\subsection{Data}
We collected short image bursts of street scenes from Google Street View. These bursts are captured in major US cities, and encompass a range of outdoor urban scenes including streets, parks, and parking lots, under lighting conditions ranging from clear to cloudy. In total, we collected $142,778$ image/background pairs for training and $16,034$ as a test set. The test images are drawn from cities near the training ones to ensure no overlap between training and test sets. We center-crop all images to $512 \times 512$ with a field of view of 75$^\circ$. To better evaluate performance, we also randomly picked a small subset ($\sim$150) of sunny images where objects cast hard shadows, and a subset ($\sim$180) of cloudy images where objects cast subtle soft shadows from the test set.

\begin{figure}[t!]
    \centering
    \includegraphics[width=\linewidth]{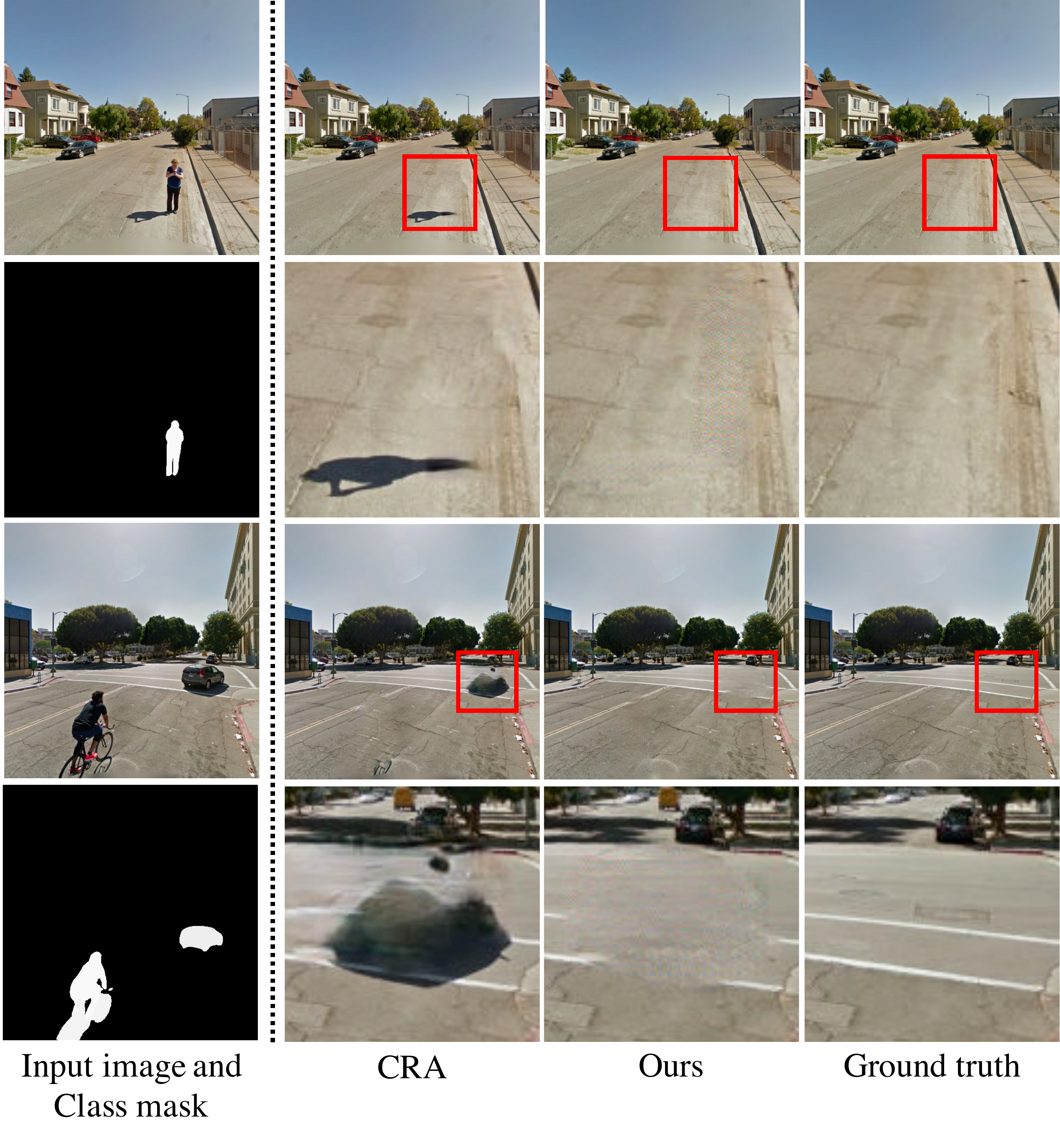}
    \caption{Object removal results on the test set. The traditional inpainting method~\cite{yi2020contextual} only inpaints the area within the mask and has leftover shadows. Our method removes objects completely along with shadows. In addition, the inpainting method fails to inpaint for large object (car in the second example).}
    \label{fig:remove_results}
    \vspace{-12pt}
\end{figure}
\begin{figure*}[t!]
    \centering
    \includegraphics[width=\textwidth]{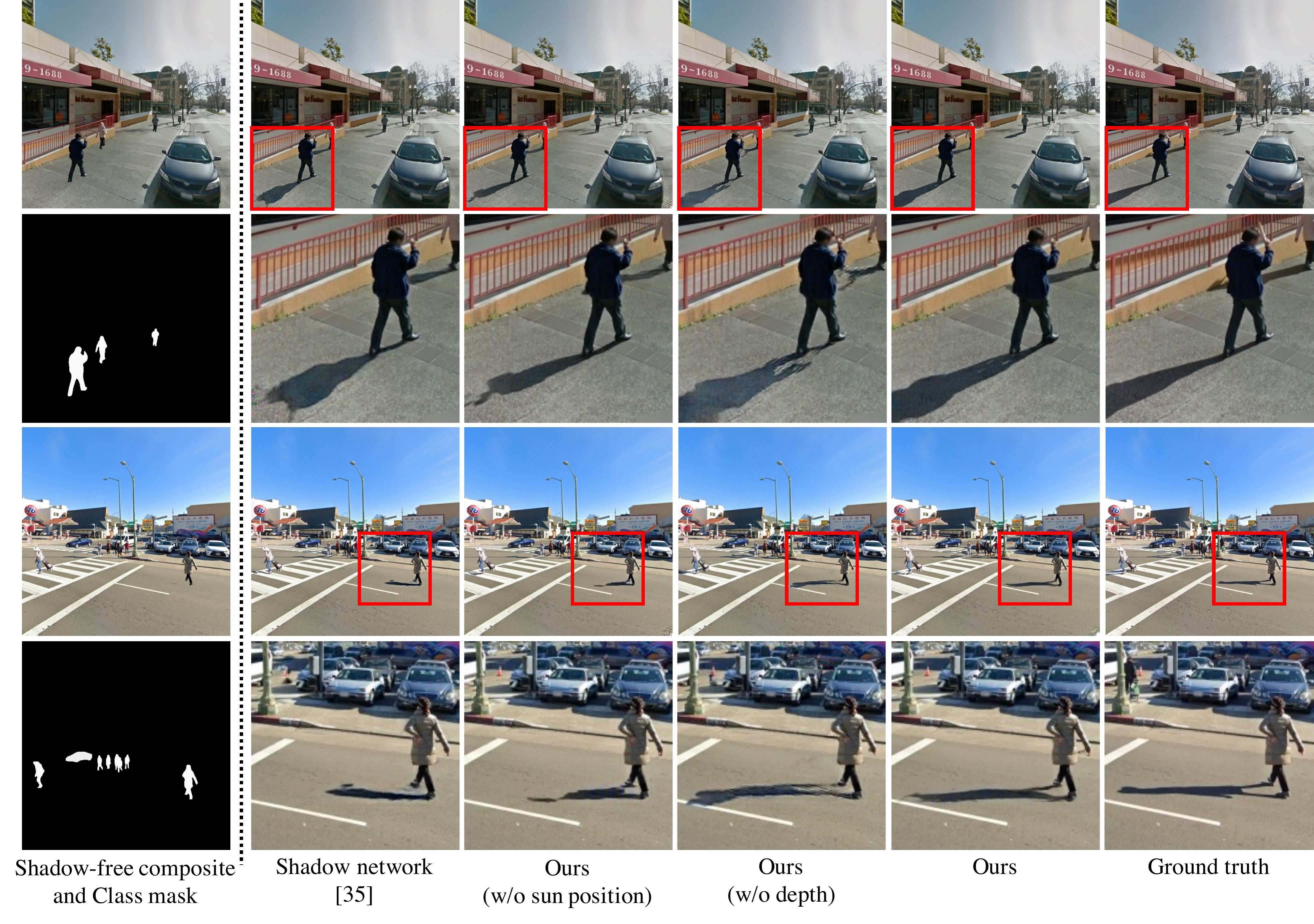}
    \caption{Object insertion results on the test set. Our method generates the most realistic shadows with details. The sun position input helps the network to determine the shape of the shadow. The depth map prevents the network from synthesizing broken or detached shadows.}
    \label{fig:insert_results}
    \vspace{-15pt}
\end{figure*}

\subsection{Sun Estimation Network}
We train our sun estimation network (Sec.~\ref{sec:sun_pos}) on the training set with ground truth supervision. Ground truth sun azimuth and elevation angles are calculated from each image's location, orientation, and timestamp using solar equations. Our network takes a street image and outputs two 
vectors describing distributions of azimuth and elevation angles. 
To compute a single pair of angles, we find the highest probability bin from each vector, and use the bin center as the estimated angle. 
\addtext{We compare our sun estimation network with~\cite{ma2017find,hold2019deep}, adding a fully connected layer to their method to predict the elevation angle.}
On average over the test set, our azimuth prediction has an angular error of $35.71\degree$ vs.\  $50.17\degree$~\cite{hold2019deep} vs.\  $52.59\degree$~\cite{ma2017find}, and our elevation prediction has an angular error of $9.79\degree$ vs.\  $13.02\degree$~\cite{hold2019deep} vs. \ $13.82\degree$~\cite{ma2017find}. We further convert the sun angles to directions on the unit sphere and compute the angle between predicted and ground truth vectors, yielding an average error
of $27.00\degree$. These error rates are reasonably low, and of suitable accuracy for applications like lighting matching and shadow prediction. 

\subsection{Removal Network}
We trained our removal network (Sec.~\ref{sec:remove}) on the training set with median images as supervision for the object removed images. And the supervision for inpainting masks is computed by thresholding the color difference between median images and original images.
Fig.~\ref{fig:remove_results} shows example object removal results on our test set using (1) the state-of-the-art inpainting method CRA~\cite{yi2020contextual} (trained on Places2~\cite{zhou2017places}), which only inpaints the mask area and (2) our method, which also predicts an enlarged inpainting mask. 
Both networks realistically inpaint the object region; however, CRA fails to remove object shadows since they are not included in the mask. Our network yields a complete object removal, which overall is more realistic. 
We show quantitative results using the LPIPS~\cite{zhang2018unreasonable} error metric in Tab.~\ref{tbl:removal}. Both methods achieve a lower error compared to the input image. Ours has much lower LPIPS error than CRA on sunny days (where our method benefits from removing hard shadows), and slightly lower on cloudy days (where the shadows are more subtle).
As shown in Fig.~\ref{fig:remove_results}, our method removes objects completely and performs better in the task of object removal.
\addtext{We further tried running CRA~\cite{yi2020contextual} using our thresholded inpainting mask. This method gave an LPIPS score of $0.162$ on the test set (vs. ours at $0.104$). CRA is not trained with our inpainting mask, thus cannot adapt errors in our mask estimation, leading to artifacts in the final output.}

\begin{table}[t!]
    \centering
    \begin{tabular}{lccc}
    \toprule
    Method  & All & Sunny & Cloudy \\
    \midrule
    Input                 & 0.113          & 0.099         & 0.096 \\
    CRA~\cite{yi2020contextual} & 0.107          & 0.090         & 0.083 \\
    Ours                        & \textbf{0.104} & \textbf{0.079} & \textbf{0.080} \\
    \bottomrule
    \end{tabular}
    \vspace{3pt}
    \caption{Object removal results on all test images, the sunny subset, and the cloudy subset, measured in LPIPS~\cite{zhang2018unreasonable}. Lower is better.}
    \label{tbl:removal}
    \vspace{-5pt}
\end{table}

\begin{table}[t!]
    \centering
    \begin{tabular}{lccc}
    \toprule
    Method  & All & Sunny & Cloudy \\
    \midrule
    Shadow-free composite  & 0.073                     &  0.060             &  0.058          \\
    Shadow network         & 0.069                     &  0.054              &  \textbf{0.056} \\
    Ours (w/o $x$-$y$ grid) & 0.079                     &  0.065              &  0.069          \\
    Ours (w/o sun position) & \textbf{0.068}            &  0.054              &  \textbf{0.056} \\
    Ours (w/o depth map)   & 0.078                     &  0.060              &  0.062          \\
    Ours                    & \textbf{0.068}            &  \textbf{0.053}     &  \textbf{0.056} \\
    \bottomrule
    \end{tabular}
    \vspace{3pt}
    \caption{Object insertion results on all test images, the sunny subset, and the cloudy subset, measured in LPIPS~\cite{zhang2018unreasonable}. Lower is better.}
    \label{tbl:insert}
    \vspace{-12pt}
\end{table}

\subsection{Insertion Network}
Our insertion network (Sec.~\ref{sec:insert}) is trained to take shadow-free composite images and 
render object shadows, using original images as ground truth supervision. Fig.~\ref{fig:insert_results} shows example results 
using (1) a baseline pix2pix-style method~\cite{wang2020people} that takes a shadow-free image and an $x$-$y$ grid; (2) an ablative method that takes a shadow-free image, $x$-$y$ grid and depth map; (3) an ablative method that takes a shadow-free image, $x$-$y$ grid, and predicted sun position; and (4) our method. All methods are trained on our training set. The predicted sun position helps the network produce shadows in the right direction. The depth map and $x$-$y$ grid stabilize training, preventing the network from overfitting and producing broken \addtext{or detached} shadows. Quantitative results shown in Tab.~\ref{tbl:insert} suggest that our method has an advantage over other models. On sunny days, our full model benefits from the depth map, sun position and $x$-$y$ grid, and outputs realistic, detailed shadows. On cloudy days, our model synthesizes subtle soft shadows, still performing the best overall.

\begin{figure}[t!]
    \centering
    \includegraphics[width=\linewidth]{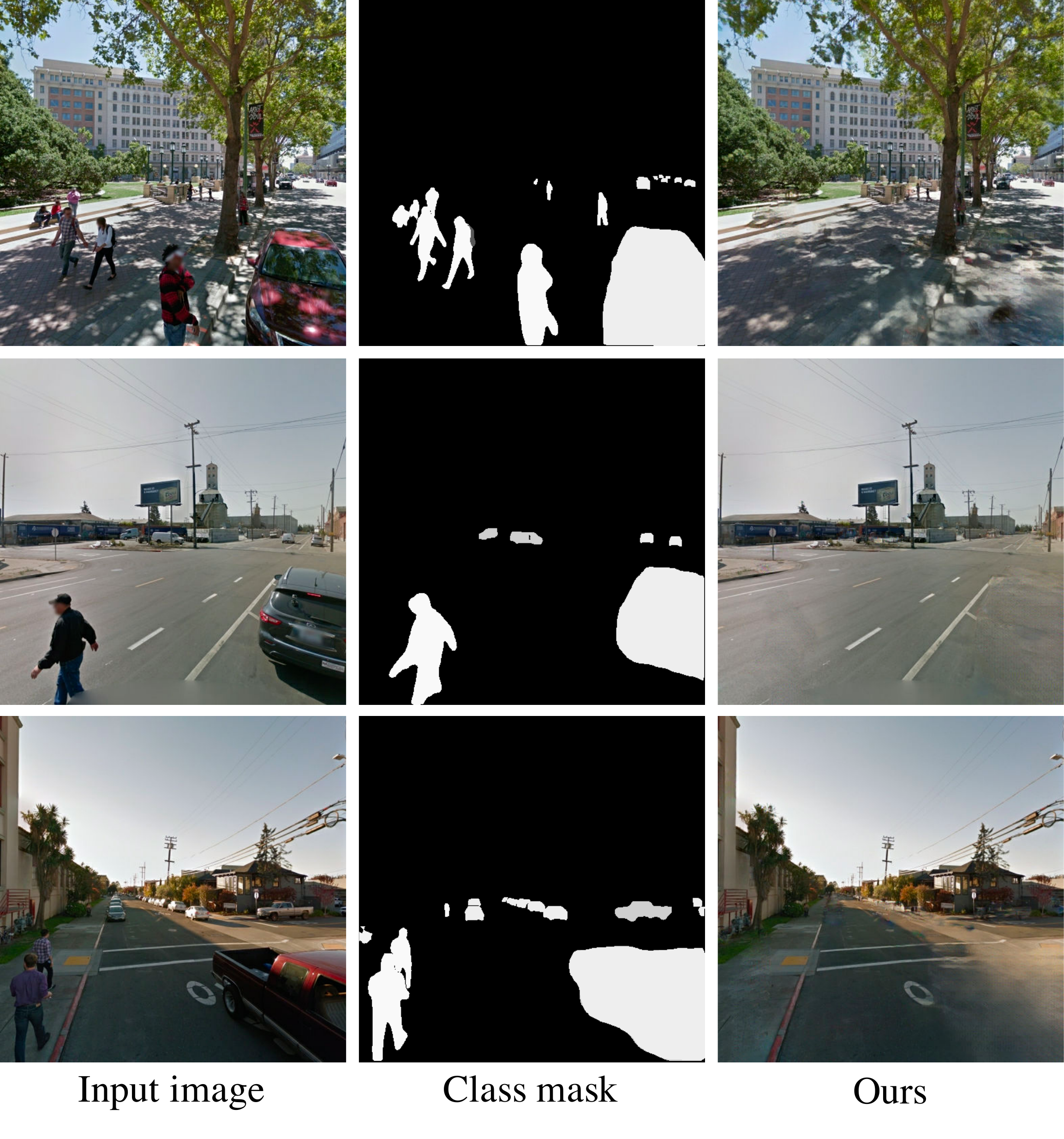}
    \caption{Qualitative results for removing all people and cars in a street image. From left to right: the input image, the class mask for objects to be removed, and the removal results generated by the removal network. Our method removes objects completely.}
    \label{fig:empty}
    \vspace{-15pt}
\end{figure}
\begin{figure}[t!]
    \centering
    \includegraphics[width=\linewidth]{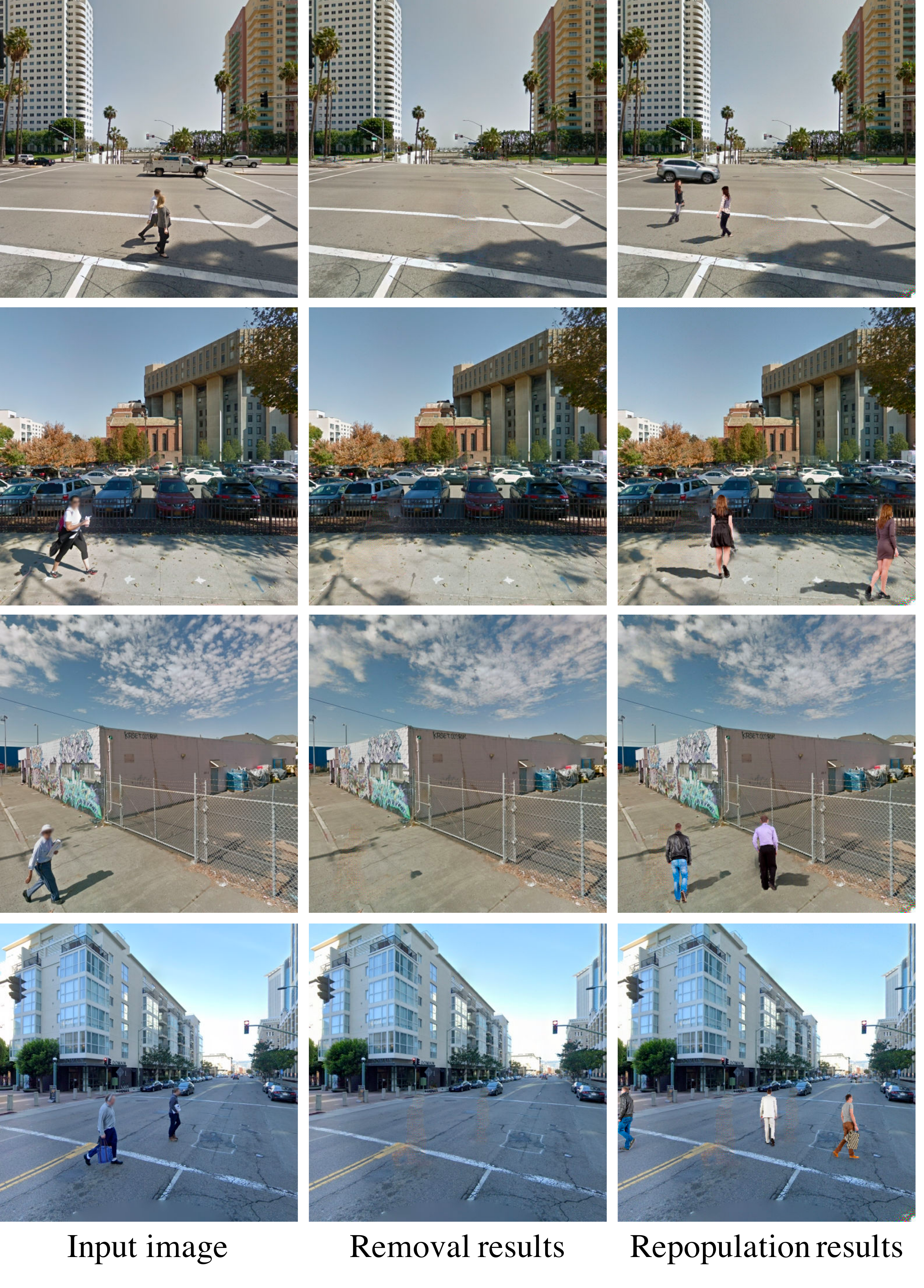}
    \caption{Qualitative results for repopulating street scenes. From left to right: the input image, the background image after removing all people, and repopulation results.
    Our pipeline selects people matching the scene's lighting, places them randomly on sidewalks and roads, and synthesizes realistic shadows.}
    \label{fig:repopulate}
    \vspace{-17pt}
\end{figure}
\section{Applications}
In this section, we discuss potential applications of our method to recomposing or repopulating single street images. These applications are enabled by one or more of the components of our pipeline. We also discuss ethical considerations involved in such applications in Sec.~\ref{sec:broader}.

\vspace{2pt} 
\noindent \textbf{Object lighting matching.}
When repopulating scenes, selecting objects with similar lighting as the scene is crucial for realistic composition. 
Hence, we wish to 
compute the sun position for both the source object and target scene. Hence, we train two sun estimation networks (Sec.~\ref{sec:sun_pos}), one for scenes and one for objects. 
The scene sun estimation network takes the image $I$ and predicts sun azimuth and elevation vectors $a_{\rm{scene}}, e_{\rm{scene}}$, while the object sun estimation network pre-computes sun angle vectors $a_{\rm{obj}, i}, e_{\rm{obj}, i}$ for each object $o_i$ in the collection. The object $o_i$ that maximizes $a_{\rm{scene}} \cdot a_{\rm{obj}, i} + e_{\rm{scene}} \cdot e_{\rm{obj}, i}$ is then selected as the object that best matches the scene's lighting. 

\vspace{0pt}
\noindent \textbf{Emptying the city.}
Mask R-CNN \cite{wu2019detectron2} can segment out certain set of objects (people, cars, bikes, etc). Our removal network then takes this mask, and synthesizes an image without those objects along with their shadows. 
This enables applications such as removing all people and cars in NYC or LA. 
As demonstrated in Fig.~\ref{fig:empty}, we use our removal network to remove all the objects---people and cars---in the image, giving users a different visualization of a city. Hence, it can also enhance the privacy of the imagery. Our method successfully removes all objects along with their shadows from the given street image. 

\vspace{0pt}
\noindent \textbf{Privacy enhancement.}
While removing all the people in the image enhances privacy, it decreases the liveliness of the street scene as well. To that end, we built a collection of people viewed from the back (or nearly the back) from licensed imagery on Shutterstock. Our pipeline can populate scenes with such people, thus enhancing privacy while retaining a sense of liveliness within the scene. 


As above, our method can remove whole categories of objects to yield a background frame $I_{\rm{back}}$. Then, we can use our object lighting matching method to find a set of best matching objects, 
then randomly place each object $o_i$ on sidewalk and road regions in 
$I_{\rm{back}}$ via the segmentation map in Sec.~\ref{sec:geometry}. Objects will be automatically resized and occluded using the methods described in Sec.~\ref{sec:geometry} to get the shadow-free composition $I_{\rm{comp}}$. 
Finally 
$I_{\rm{comp}}$ 
is passed to the insertion network to synthesize the final composition $I_{\rm{final}}$. 
Fig.~\ref{fig:repopulate} shows results for repopulating street scenes. We substitute the people in the scene with anonymized people, thus enhancing privacy while preserving the realism of street scenes.
Note that our work focuses on lighting, and 
does not attempt to match the camera viewpoint for inserted objects as in~\cite{lalonde2007photo} \addtext{or compensate for differences in camera exposure, white balance, etc. These are left as future work.}

\vspace{2pt}
\noindent \textbf{Other applications.}
We have also developed an interactive scene reconfigurator that leverages the elements of our framework. With this tool, a user can take a street image and remove selected existing objects, or conversely, place new objects in the scene. This tool can synthesize street images that are rare in real life, e.g., people walking in the middle of a busy road, or cars driving on the sidewalk. These synthesized scenes could be used for data augmentation for autonomous driving to simulate dangerous situations.


\section{Discussion and Ethical Considerations}
\label{sec:broader}
In this paper, we introduced a fully automatic pipeline for populating, depopulating, or repopulating street scenes. Our pipeline consists of four major components: (1) a removal network that can remove selected objects along with their shadows; (2) a sun estimation network that predicts the sun position from an image; (3) a method to scale and occlude inserted objects properly; and (4) an insertion network that synthesizes shadows for inserted objects. These components are trained on short image bursts of street scenes, and can run on a single street image at test time.  Further, we show multiple applications of our pipeline for depopulating and repopulating street scenes.

While our work is motivated by goals like improving visualizations of scenes and enhance image privacy, it is important to consider the broader impacts and ethical aspects of computer vision research, particular work related to synthetic imagery.
Potential harmful outcomes relating to recomposing street scenes include (1) misuse in creating a false narrative, such as a crowd or protest in a certain location, and (2) misrepresenting a neighborhood by changing the demographics of people therein.
In our case, some issues related to synthetic media are mitigated by inherent limitations of our method---for instance, our method can compose separated people into scenes, and synthesize their shadows cast on the ground, but would have trouble generating a dense crowd of people where people would be shadowing each other.
That said, any deployment of our methods in a real-world setting would need careful attention to responsible design decisions. Such considerations could include clearly watermarking any user-facing image that has been recomposed, and matching the distribution of anonymized people composed into a scene to the underlying demographics of that location. 
At the same time, our work may lead to knowledge useful to counter-abuse teams working on manipulated imagery and synthetic media data methods.

\newpage

{\small
\bibliographystyle{ieee_fullname}
\bibliography{refs}
}

\end{document}